# Automatic annotation of visual deep neural networks

Ming Li[1,2], ChenHao Guo[1,2*]

(1. College of Mathematics and Computer Science, Fuzhou University, Fuzhou, 350108, China
2. Fujian Provincial Key Laboratory of Network Computing and Intelligent Information Processing, Fuzhou, 350108, China)

**Abstract:** Computer vision is widely used in the fields of driverless, face recognition and 3D reconstruction as a technology to help or replace human eye perception images or multidimensional data through computers. Nowadays, with the development and application of deep neural networks, the models of deep neural networks proposed for computer vision are becoming more and more abundant, and developers will use the already trained models on the way to solve problems, and need to consult the relevant documents to understand the use of the model. The class model, which creates the need to quickly and accurately find the relevant models that you need. The automatic annotation method of visual depth neural network proposed in this paper is based on natural language processing technology such as semantic analysis, which realizes automatic labeling of model application fields. In the three top international conferences on computer vision: ICCV, CVPR and ECCV, the average correct rate of application of the papers of 72 papers reached 90%, indicating the effectiveness of the automatic labeling system.

## 0. introduction

Vision is an important way for human beings to understand the external world. In the process of human cognition, more than 80% of the information comes from the visual system [1]. As an important research direction in the field of computer science, computer vision aims to enable computers to help or replace human eyes to perceive images, videos or multidimensional data, and obtain target information and data [2]. With the development of computer vision, various research directions have been explored, such as target detection, target tracking, super-resolution correlation, image generation, 3D modeling, human posture correlation, image restoration, image classification, image edge, image segmentation, feature detection and matching, and so on.

Nowadays, with the further research on deep neural network, it has also been widely used in the field of computer vision, especially in target detection, target tracking, super-resolution, image generation, 3D modeling and human posture correlation [3]. Therefore, the complex model with more hidden layers is proposed, which has more powerful and effective feature learning and feature expression ability than the traditional machine learning methods. Facing more and more models, developers have the need to find the appropriate model quickly and accurately according to the problems they encounter. In view of this demand, this paper proposes an automatic annotation system in the model application field. Through the automatic annotation of the model, it can help developers understand the application field of the model more quickly and accurately, so as to judge whether it is the model they need.

At present, developers learn to understand the relevant models by reading the paper documents, and the abstract of the computer vision class of the paper often contains the key information that can express the main work and contribution of the paper. Therefore, according to the abstracts of computer vision papers, this paper obtains the key information we want to extract through word segmentation and syntactic analysis of natural language processing, and then realizes the automatic annotation of the application field of the model. The experimental results show that high accuracy is achieved in the judgment of the application fields corresponding to the model described in the paper documents, which proves that this idea is practical and effective.

The main contributions of this paper are as follows:

1) The visual deep neural network is divided into six fields, namely target detection, target tracking, super-resolution, image generation, 3D modeling and human posture correlation. The papers collected on the network are used as the experimental corpus to select the keywords of the whole abstract without using the extraction model, Finally, the architecture diagram of visual depth neural network is constructed. According to the architecture diagram, we can intuitively see the proportion of different keywords in different fields.

2) By observing the abstract characteristics of visual neural network papers, it is found that keywords, as the expression of phrases that can best reflect the characteristics of the field to which the model belongs, often have similarity and fixity in the places in the text. According to this finding, eight models that can extract the keyword groups of article abstracts are proposed.

3) According to the constructed architecture diagram, the similarity between the extracted key phrases and six fields is calculated, and the field with the highest similarity is identified as the application field of the model. The experimental results show that the average accuracy of the discrimination method can reach more than 90%.

Section 1 summarizes the relevant research of experts and scholars at home and abroad in natural language processing and text classification technology; Section 2 introduces the overall framework of this work, and introduces the components of the framework in detail; Section 3 introduces the experimental data and result analysis; Finally, it summarizes the full text and analyzes the direction that can continue to move forward in the future.

# 1. Related work

**1.1 Natural Language Processing**

The idea of natural language processing has gradually attracted people's attention since Turing put forward the famous "Turing test" [4] in 1950.

After that, until the 1970s, natural language processing mainly adopted the rule-based method, that is, the term extraction was completed through the term dictionary and Rule Template written by experts [5,6,7]. However, the rule-based method has inevitable disadvantages. Firstly, the rules cannot cover all sentences. Secondly, this method is extremely demanding for developers. Developers should not only be proficient in computers, but also proficient in linguistics. Therefore, although some simple problems have been solved at this stage, they can not fundamentally put natural language understanding into practice.

After the 1970s, based on statistical methods, that is, based on the assumption of high adhesion between words within terms, term extraction was realized by using statistical features [8]. At present, statistical features include chi square test, log likelihood test, mutual information [9] and C-value / ncvalue [10]. However, the effect of relying solely on internal adhesion is not ideal. In order to greatly improve the accuracy, gasterson and Katz used a part of speech filter to filter candidate phrases in 1995, which only allows possible "phrase" patterns to pass through [11]. In addition, statistical methods also have some shortcomings, such as mutual information algorithm is difficult to eliminate the interference of ultra-low frequency words and ultra-high frequency words in the corpus.

Since 2008, experts and scholars have also shifted their thinking to apply deep learning to natural language processing research, from the initial word vector [12] to the word2vec algorithm created by the research team led by Thomas mikelov in Google in 2015 [13]. The combination of deep learning and natural language processing has reached a climax, and has achieved some success in the fields of machine translation, reading comprehension and so on. Nowadays, RNN has been one of the most commonly used methods in natural language processing. Gru [14], LSTM [15] and other models have triggered waves of upsurge one after another.

**1.2 Text classification**

In the development of text classification, many methods and ideas have been put forward [16]. The following lists six innovative methods in this process, and describes the advantages and disadvantages of different methods.

1) Word bag model: a technology used for text information retrieval and text classification in the field of natural language processing, which is very easy to understand and implement, and provides great flexibility for customizing specific text data [17]. However, discarding word order ignores the context, which in turn affects the semantics of words in the document.

2) TF-IDF [18]: although important words can be found, these important words are discrete, which will lead to the neglect of the commonness between important words.

3) Word embedding [19]: Although the effect of word embedding method is even better than CNN / RNN in some text classification tasks, if a sentence is very long, a small amount of important information will be submerged by most useless information.

4) Textcnn / textrnn [20]: compared with word embedding, it is more suitable for the analysis of long text. Textcnn is good at capturing shorter sequence information and textrnn is good at capturing longer sequence information, but they are difficult to capture the correlation between long-term context information and discontinuous words.

5) Attention mechanism [21]: it can help the model give different weights to each part of the input, extract more critical and important information, and make the model make more accurate judgment without bringing greater overhead to the calculation and storage of the model.

6) Matchpyramid [22]: use the similarity calculation to construct the similarity matrix of the text, and then convolute to extract the features. The text matching is processed into image recognition.

# 2. Method design

**2.1 Overall Structure**

The automatic extraction process of visual deep neural network label information is shown in Figure 1.

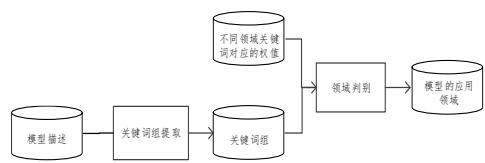

Fig. 1 Overall Framework

Firstly, the model description is taken as the input, the keywords of the article can be obtained according to the

keyword extraction, and then the application field of the model is calculated according to the obtained keywords and the corresponding weights of keywords in different fields.

**2.2 Discrimination of model research field**
**2.1.1 Architecture design of visual depth neural network**

At present, deep neural network is widely used in many fields of computer vision, such as graphics classification, object detection, pose estimation, image segmentation and face recognition [23]. This paper mainly studies six of them. Hundreds of papers in these six fields are collected as experimental corpus from the computer vision papers published on the network. Abstracts containing various fields are randomly selected from the experimental corpus. Without extraction model extraction, the keywords of various categories are selected for the whole abstract to form the assignment word set of each field.

After continuous experiments, the architecture diagram of visual deep neural network is finally constructed, as shown in Figure 2. The larger the word size in the third layer represents the greater proportion of the word in this field, while the words with the same color represent the same proportion. The fourth layer is the specific weight of the corresponding keywords in each field in the third layer. Here, only the first five are listed in each field.

**2.2.2 Extraction of model keywords**

By observing the abstract characteristics of visual neural network papers, it is found that keywords, as a phrase expression that can best reflect the characteristics of the field to which the model belongs, often have similarity and fixity in the places in the text. According to this finding, this paper proposes eight keyword extraction models. The following are the definitions and examples of the model.

1) Extraction model 1: first sentence subject

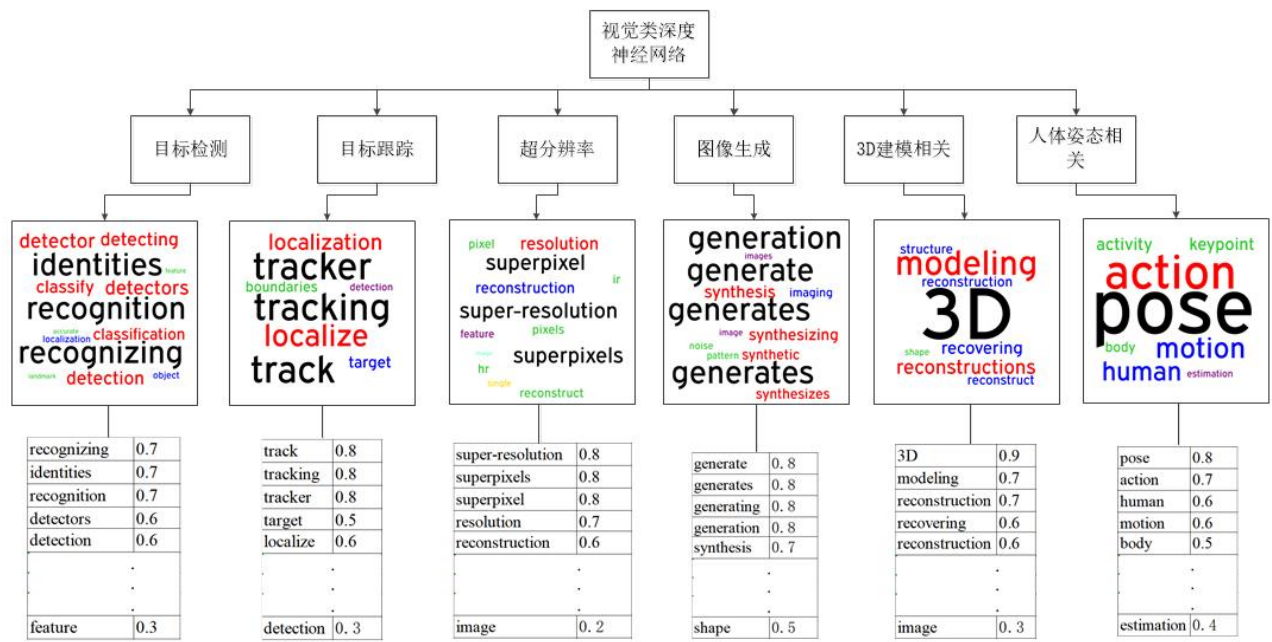

Fig. 2  Vision-like deep neural network architecture

In the abstracts of computer vision papers, the problems that the papers try to solve are generally reflected in the first sentence, and the subject of the first sentence will generally express the problem. Therefore, the phrase corresponding to the subject of the first sentence is of great reference and extraction value for the discrimination of content.

Extract the definition of model 1: starting from the root (root node) of the sentence dependency tree, backtrack to find the entity phrase with nsubj (nominal subject) dependent on the root. If it is not found, take the entity phrase with dobj (direct object) as the first found dependency.

Take sentence S1 as an example:

s1="Material recognition for real-world outdoor surfaces has become increasingly important for computer vision to support its operation in the wild".

The dependency tree corresponding to S1 is shown in Figure3:

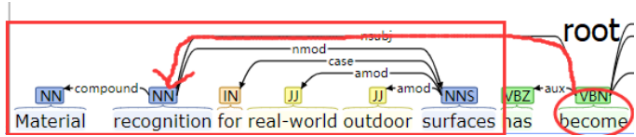

Fig. 3   Dependent analysis tree corresponding to $s_1$

According to the definition of extraction model 1, from the root, we can get the word recognition with nsubj relationship, and then take the word as the starting point to look forward and backward for the word that is dependent on the word and is a modification relationship, and extract the whole entity phrase: "material recognition for real world outdoor surfaces".

2) Extraction model 2: specific nouns and their modifiers

In the abstract of computer vision papers, the information describing the problems solved or the fields involved in the article often contains some specific nouns (as shown in Table 1). Extracting the corresponding phrases for these specific nouns can also help to identify the domain to which the model belongs.

Definition of extraction model 2: take a specific noun as the nominal subject and trigger condition of an entity phrase, backtrack the modifiers that are dependent on the nominal subject, traverse backward at the same time, and extract the modifications involved in the case (adverbial) relationship and conj (connective) as suffix modification information.

Take sentence S2 as an example:

s2="Deep networks have produced significant gains for various visual recognition problems."

The dependency tree corresponding to S2 is shown in Figure 4:

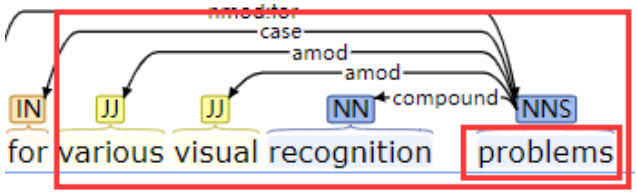

Fig. 4   Dependent analysis tree corresponding to $s_2$

According to the definition of extraction model 2, the problems in this sentence are specific nouns, and the modifiers dependent on the modified words are obtained by backtracking. Finally, the phrase "variable visual recognition problems" is extracted. Since the problems have no suffix, there is no need to traverse backward.

3) Extraction model 3: case / mark pointing part of specific nouns

According to the observation, the phrases referring to the main work of the paper often appear in the part pointed to by the case (adverbial) or mark (mainly in the part with "that", "whether" or "because") of a specific noun (as shown in Table 1). Based on this phenomenon, three extraction models are proposed.

Definition of extraction model 3: starting from a specific noun, check whether there is case / mark relationship guidance that limits the scene, field and application scope. If so, according to the different guidance rules of case and mark: for case, the entity phrase pointed to by case is extracted directly; For mark, we traverse backward to find the word whose dependency is dobj (direct object), and then judge whether the word between mark and dobj is dependent on mark or the word pointed to by mark. If so, it is added to the pre modification of the last extracted phrase.

Take sentence S3 as an example:

s3="This paper addresses the problem of estimating and tracking human body keypoints in complex multi-person video."

The dependency tree corresponding to S3 is shown in Figure 5:

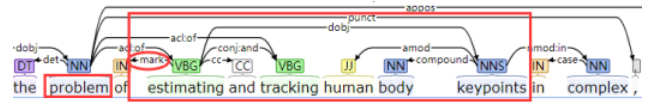

Fig. 5   Dependent analysis tree corresponding to $s_3$

According to the definition of extraction model 3, the problem of in this sentence is a specific noun, followed by Mark's guidance, so look back for the direct object dependent on mark to get keypoints, then extract the whole entity phrase, and finally get "estimating and tracking human body keypoints".

4) Extraction model 4: direct object of specific transitive verbs

In the abstracts of computer vision papers, specific transitive verbs (as shown in Table 1) often have direct objects that can often represent the main work of the article, so such direct objects are the target phrases to be extracted. For this rule, an extraction model four is proposed.

Extract the definition of model 4: detect the keyword in the sentence, then find the direct object of the keyword, that is, the word with dobj (direct object), and obtain the entity phrase corresponding to the word according to the index.

Take sentence S4 as an example:

s4="We present approach to improve the precision of facial landmark detectors on images."

The dependency tree corresponding to S4 is shown in Figure 6:

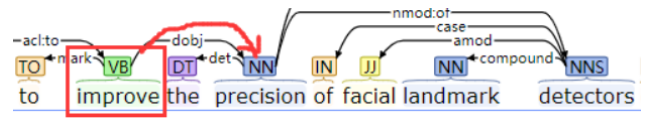

Fig. 6   Dependent analysis tree corresponding to $s_4$

According to the definition of extraction model 4, improve in this sentence is a specific transitive verb, so its direct object precision is extracted, and then the entity phrase is extracted according to the precision to obtain "precision of facial landmark detectors".

5) Extraction model 5: the verb phrase corresponding to the direct object of a specific verb

For a specific verb (as shown in Table 1), the content expressing the main article work often appears in the clause of the modified object, while the object modified by a specific verb usually acts as a subject in the clause, and the corresponding verb phrase is the target phrase.

Extract the definition of model 5: firstly, detect the word whose dependency is dobj (direct object) corresponding to the specific verb. If there is clause information later, continue to traverse backward to find the verb dependent on the direct object, and obtain the verb phrase as the target phrase according to the found verb.

Take sentence S5 as an example:

s5= "We propose a deep but compact convolutional network to directly reconstruct the high resolution image."

The dependency tree corresponding to S5 is shown in Figure 7:

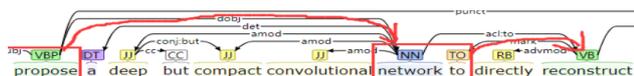

Fig. 7　Dependent analysis tree corresponding to $s_5$

According to the definition of extraction model 5, propose is a specific verb in this sentence, and then find the corresponding direct object network. Since there is clause information after it, find the verb corresponding to network, namely reconstruct, and finally find the verb phrase corresponding to reconstruct "reconstruct the high resolution image".

6) Extraction model 6: for guided phrases

In the abstract, for often appears at the beginning of the sentence. As the limitation of the problem to be solved, it plays a certain role in indicating the research direction of which computer vision category the paper belongs to. At the same time, due to different sentence patterns and expression methods, for guidance is also used in the sentence. Based on the above observations, the extraction model 6 is designed and implemented.

Definition of extraction model 6: first, judge whether the first sentence is for and the dependency is case (adverbial). If it is satisfied, extract the entity phrase according to the object indicated by the dependency of case. If the beginning of the sentence is not for, traverse the sentence to find the word for, and extract the target phrase when the same conditions above are met.

Take sentence S6 as an example:

s6="For modeling the 3D world behind 2D images, a polygon mesh is a promising candidate for 3D representation."

The dependency tree corresponding to S6 is shown in Figure 8:

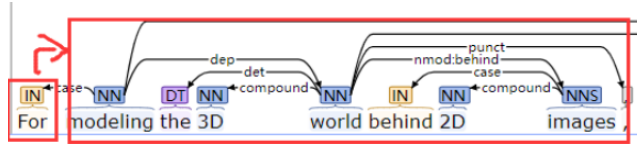

Fig. 8　Dependent analysis tree corresponding to $s_6$

According to the definition of extraction model 6, the sentence satisfies the condition that the first sentence is for and the dependency is case, so find the modeling dependent on for and the dependency is case. Finally, extract the entity phrase according to the modeling to obtain "modeling the 3D world behind 2D images".

7) Extraction model 7: objects guided by specific transitive verbs

In the abstracts of computer vision papers, there are generally sentences in which the subject is we and there are specific transitive verbs (as shown in Table 1) to express key information. Based on this observation, the extraction model 7 is designed.

Extract the definition of model 7: if the subject of the sentence is we, find the specific transitive verb in the sentence, continue to traverse backward after finding the specific transitive verb, find the word dependent on the transitive verb, extract the pre modification and suffix modification according to the word, and sort and combine them as the target phrase.

Take sentence S7 as an example:

s7= "We propose a novel visual tracking algorithm based on the representations from a discriminatively trained Convolutional Neural network."

The dependency tree corresponding to S7 is shown in Figure 9:

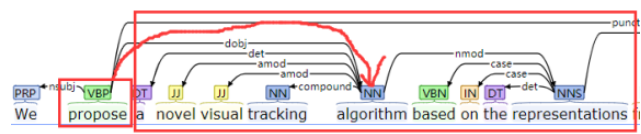

Fig. 9　Dependent analysis tree corresponding to $s_7$

According to the definition of extraction model 7, the subject of this sentence is we, and propose is a specific transitive verb. Continue to traverse backward to get the word algorithm dependent on propose, and then extract the pre modification and suffix modification according to the algorithm to finally get the most target phrase of "a novel visual tracking algorithm based on the representations".

8) Extraction model 8: objects guided by specific intransitive verbs

Similarly to extraction model 7, there are sentences with the subject we and specific intransitive verbs (as shown in Table 1) to express key information. Based on this observation, extraction model 8 is designed.

Extract the definition of model 8: if the subject of the sentence is we, find the specific intransitive verb in the sentence. After finding the intransitive verb, continue to traverse backward to find the object that is dependent on the word. Extract the pre modification and suffix modification according to the object, and sort and combine them as the target phrase.

Take sentence S8 as an example:

s8 ="We focus on the task of amodal 3D object detection in RGB-D images."

The dependency tree corresponding to S8 is shown in Figure 10:

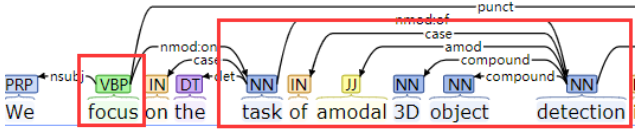

Fig. 10 Dependent analysis tree corresponding to $s_8$

According to the definition of extraction model 8, the subject of this sentence is we, and focus is a specific non transitive verb. Continue to traverse backward to get the object task dependent on focus, and then extract the pre modification and suffix modification according to the task to finally get the most target phrase of "task of amodal 3D object detection".

Because the abstracted extraction models are different, it is necessary to combine different specific words to complete the extraction of target phrases. The specific words corresponding to the extraction model are shown in Table 1:

Tab. 1 Specific words corresponding to the extract model

| Extraction model | key word |
| --- | --- |
| Specific nouns and their modifiers | ['problem','task','system','algorithm','techniqu','method'] |
| The case / mark pointing part of a specific NOUN | ['sucess','result','progress','problem','attention','approach','tool','advance','technique','method','framework','algorithm','research','role','benchmark','system','Network','network','datset','task','improvement','moel','promise','performance','state-of-the-art'] |
| Direct object of specific transitive verbs | ['help','improve','disentangle','stud'] |
| A verb phrase corresponding to the direct object of a particular verb | ['propose','present','introduce','identif','facilitate','show'] |
| Object guided by specific transitive verbs | ['propose','present','introduce','identif','facilitate','show'] |
| An object or verb phrase guided by a specific intransitive verb | ['focus','learn','rely','aim'] |

According to the above eight models, the key phrases of the summary are extracted. In order to facilitate subsequent formula calculation and application domain discrimination, the extracted target phrases need to be transformed into a set of words. Here, the idea of word bag model [24] is adopted.

$$patternRseult = getPattern\operatorname{Re}sult(sent) \quad (1)$$

$$words_i = getWords(pattern\operatorname{Re}sult_i) \quad (2)$$

$$wordBag = set(words_i) \quad (3)$$

Where (1) is the target phrase extracted by all extraction models combined with keyword matching; In formula (2), the phrase obtained by the matching of the second model is decomposed into word sets in units of words; Equation (3) is the set of all contained words。

### 2.2.3 Domain discrimination

According to the keyword groups extracted from the eight models in 2.2.2 above and the visual depth neural network architecture constructed in 2.2.1, the domain to which the model belongs can be calculated through the discrimination formula. The discrimination publicity here uses cosine similarity [25] to judge which domain the prediction model matches better. The specific steps are as follows:

Step 1: calculate the domain vectors of six domains according to formula (4) and the weights corresponding to the keywords in the fourth layer of the visual depth neural network architecture constructed in 2.2.1. Formula (4) is the weight corresponding to the j-th keyword in the field. There are m keywords in total (the value of M is determined according to the number of keywords in the field). Finally, the field vectors of six fields are obtained, as shown in Figure 11.

$$x_i = (w_{i1}, w_{i2}, ..., w_{ij}, ..., w_{im})^T, i=1,...,6 \quad (4)$$

Step 2: calculate the model domain vector of the model in six domains according to formula (5) and the keyword set extracted from the eight models in 2.2.2. Formula (5) is the model weight of the j-th word in the domain. There are two cases for the value of. If the keyword set extracted by the model contains the j-th word in the I domain, it is consistent with the value, otherwise it is 0.

$$y_i = (z_{i1}, z_{i2}, ..., z_{ij}, ..., z_{im})^T, i=1,...,6 \quad (5)$$

Step 3: calculate the matching degree between the model and the six fields. The sum vector can be obtained from step 1 and step 2, and then calculate the cosine similarity through formula (6), indicating the similarity of the model in the second field. The larger the value, the closer it is to 1, indicating the better matching with this field.

$$sim_i = \cos(x_i, y_i) = \frac{\sum_{j=1}^{m} w_{ij} z_{ij}}{\sqrt{\sum_{j=1}^{m} w_{ij}^2} \sqrt{\sum_{j=1}^{m} z_{ij}^2}}, i=1,...,6 \quad (6)$$

Step 4: calculate the k-th matching field of the model. According to formula (7), the field close to the model in the six fields can be obtained. When k is 1, it indicates the application field to which the maximum probability of the model belongs.

$$top_k = MAX(sim_{1,2,...,6})_k, k=1,...,6 \quad (7)$$

## 3. Experiment

### 3.1 experimental data

| 领域 | (关键词，权值) | 领域向量 |
|---|---|---|
| 目标检测 | (recognizing, 0.7), (identities, 0.7), (recognition, 0.7), (classify, 0.6), (detector, 0.6), (detectors, 0.6), (detection, 0.6), (detecting, 0.6), (classification, 0.6), (object, 0.5), (localization, 0.5), (accurate, 0.3), (landmark, 0.3), (feature, 0.3) | $x_1 = (0.7, 0.7, 0.7, 0.6, 0.6, 0.6, 0.6, 0.6, 0.6, 0.5, 0.5, 0.3, 0.3, 0.3)^T$ |
| 目标跟踪 | (track, 0.8), (tracker, 0.8), (tracking, 0.8), (localize, 0.6), (localization, 0.6), (target, 0.5), (boundaries, 0.4), (detection, 0.3) | $x_2 = (0.8, 0.8, 0.8, 0.6, 0.6, 0.5, 0.4, 0.3)^T$ |
| 超分辨率 | (superpixels, 0.8), (superpixel, 0.8), (super-resolution, 0.8), (resolution, 0.7), (reconstruction, 0.6), (reconstruct, 0.5), (HR, 0.5), (LR, 0.5), (pixel, 0.5), (pixels, 0.5), (feature, 0.4), (single, 0.3), (image, 0.2) | $x_3 = (0.8, 0.8, 0.8, 0.7, 0.6, 0.5, 0.5, 0.5, 0.5, 0.5, 0.4, 0.3, 0.2)^T$ |
| 图片生成 | (generation, 0.8), (generate, 0.8), (generates, 0.8), (generating, 0.8), (synthetic, 0.7), (synthesizing, 0.7), (synthesis, 0.7), (synthesizes, 0.7), (imaging, 0.6), (pattern, 0.4), (noise, 0.4), (image, 0.3), (images, 0.3) | $x_4 = (0.8, 0.8, 0.8, 0.8, 0.7, 0.7, 0.7, 0.7, 0.6, 0.4, 0.4, 0.3, 0.3)^T$ |
| 3D建模相关 | (3D, 0.9), (modeling, 0.7), (reconstructions, 0.7), (recovering, 0.6), (reconstruction, 0.6), (reconstruct, 0.6), (structure, 0.6), (shape, 0.5) | $x_5 = (0.9, 07, 0.7, 0.6, 0.6, 0.6, 0.6, 0.5)^T$ |
| 人体姿态相关 | (pose, 0.8), (action, 0.7), (human, 0.6), (motion, 0.6), (keypoint, 0.5), (body, 0.5), (activity, 0.5), (estimation, 0.4) | $x_6 = (0.8, 0.7, 0.6, 0.6, 0.5, 0.5, 0.5, 0.4)^T$ |

Fig. 11 Field vectors corresponding to six fields

The experimental data came from three top international conferences on computer vision in recent three years: iccv, CVPR and ECCV. A total of 72 papers were collected. Among them, there are 14 articles on target detection, 11 articles on target tracking, 14 articles on super-resolution, 11 articles on image generation, 10 articles on 3D modeling and 12 articles on human posture.

**3.2 experimental design**

Firstly, the model input is subjected to sentence segmentation, word segmentation and dependency analysis, and then the word set and dependency analysis tree required for the experiment are obtained. Then the keywords of the input model are extracted through the eight extraction models in 2.2.2. Finally, the application domain discrimination results of the input model can be obtained according to the process in 2.2.3.

Taking Fig. 12 as an example, after inputting the model, the keyword extracted by the model, model domain vector and model domain discrimination results are obtained. The model application domain discrimination results show that the most matching domain is human posture correlation, followed by 3D modeling correlation and target detection, because the similarity with the other three categories is 0 (it can be seen that the value of the model domain vector in these three domains is 0), so only three matching results are output here.

**3.3 Experimental results and analysis**

The experimental results of 72 papers are shown in Figure 13. In order to deeply understand the effect of automatic model annotation, this experiment calculates the accuracy of the top two with the highest similarity in model domain prediction, that is, the total accuracy of Top1 and top2 corresponding to the output results. The blue part represents the accuracy of Top1 and the yellow part is top2

Fig. 12 Experimental example

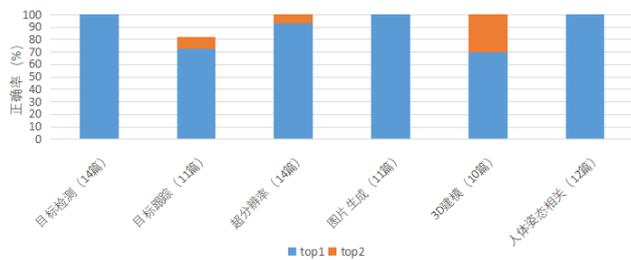

Fig. 13 Correctness of Top1 and top2

Looking at Top1 alone, the accuracy of target detection, image generation and human posture correlation can reach 100%, followed by super-resolution and target tracking, 93% and 73% respectively, and the lowest 3D modeling is 70%. When the accuracy of top2 is introduced, it can be clearly seen that except that the accuracy of target tracking is 82%, the accuracy of the other five fields can reach 100%

Based on the above results, it shows that the automatic annotation system can basically realize the accurate judgment of the input model. When using this system, developers can quickly judge whether the model is their own model by directly outputting the results of Top1 and top2.

## 4. Conclusion

In this paper, a function of automatic annotation of visual depth neural network model is proposed. Firstly, it analyzes the sentence segmentation, word segmentation and dependency of the abstract. Then, the proposed eight models are used to extract keywords, and the similarity between the model and various fields is calculated according to the constructed visual depth neural network architecture, so as to judge the application field of the model.

However, the experimental data in this paper is not rich enough, and more experimental data are needed to verify the model more accurately, which is conducive to better discover the advantages and disadvantages of the system. In the future research, we will consider adding more experimental data to improve the universality of the system.